# Towards Automated Post-Earthquake Inspections with Deep Learning-based Condition-Aware Models

Vedhus Hoskere [a,b], Yasutaka Narazaki [a], Tu A. Hoang [a], B. F. Spencer Jr. [a]

[a] Department of Civil and Environmental Engineering, University of Illinois at Urbana-Champaign, USA
[b] Department of Computer Science, University of Illinois at Urbana-Champaign, USA

## Abstract

In the aftermath of an earthquake, rapid structural inspections are required to get citizens back in to their homes and offices in a safe and timely manner. These inspections gfare typically conducted by municipal authorities through structural engineer volunteers. As manual inspections can be time consuming, laborious and dangerous, research has been underway to develop methods to help speed up and increase the automation of the entire process. Researchers typically envisage the use of unmanned aerial vehicles (UAV) for data acquisition and computer vision for data processing to extract actionable information. In this work we propose a new framework to generate vision-based condition-aware models that can serve as the basis for speeding up or automating higher level inspection decisions. The condition-aware models are generated by projecting the inference of trained deep-learning models on a set of images of a structure onto a 3D mesh model generated through multi-view stereo from the same image set. Deep fully convolutional residual networks are used for semantic segmentation of images of buildings to provide (i) damage information such as cracks and spalling (ii) contextual information such as the presence of a building and visually identifiable components like windows and doors. The proposed methodology was implemented on a damaged building that was surveyed by the authors after the Central Mexico Earthquake in September 2017 and qualitatively evaluated. Results demonstrate the promise of the proposed method towards the ultimate goal of rapid and automated post-earthquake inspections.

## 1. Introduction

Earthquakes can take a significant toll on the structural condition of buildings and thus, in the aftermath of an earthquake, structural inspections are required before allowing citizens to safely resume the occupation of these buildings. The time-consuming and laborious nature of these inspections means that citizens may have to wait for days, or even weeks, before they can return to their homes and workplaces. For example, in the 2017 central Mexico earthquake that occurred on the 19th of September 2017, the Civil Engineering Association of Mexico took about 3 weeks to complete evaluations of the entire city [1–4]. Hundreds of citizens were barred by civil protection authorities from entering any buildings with signs of damage and thus were forced to camp out in tents while they waited for these inspections. Camera-enabled unmanned aerial vehicles (UAVs) provide a promising solution to dramatically speed up manual inspections and allow for remote data acquisition from regions that may otherwise be inaccessible. Recently UAVs have begun to be used for a wide range of applications with regards to structural assessments ranging from bridge inspections (FDOT [5], MDOT[6]) to modal analysis [7,8].

For large scale applications like the inspection of an entire city, automated post-processing techniques will be vital to digest the large amount of image data that will be generated from UAV inspections. As a result, research in the use of computer vision techniques to automatically detect damage is being pursued intensely.

Recently, researchers have studied the application of deep learning algorithms, with specific focus on the use of deep convolutional neural networks (CNNs) to problems of damage identification and automated structural inspection. Yeum, et al. [9] utilized a CNN for the extraction of important regions of interest in highway truss structures to facilitate the inspection process. R-CNNs have been used for spall detection in a post disaster scenario [10], although, the results (about 59.39% true positive accuracy) leave much room for improvement. Other authors [11] have used region based deep-learning to identify concrete cracks, different levels of corrosion and steel delamination. Narazaki et al. [12] used a multi-scale deep network to identify bridge components. Hoskere, et al. [13] proposed the use of deep-learning based semantic segmentation for multiple types of damage identification using a multi-scale deep convolutional neural network. A detailed review of advances in vision based inspection and monitoring can be found in Spencer et al. [14].

In order to conduct global assessments of a structure, an important aspect to be considered is the completeness of information present. There has been limited research thus far that incorporates context of the damage and information from the entire structure to contribute to a structural assessment. Narazaki et al. [15] proposed the use of recurrent neural networks with video data to help better understand the structural component context of close up videos during bridge inspections.

In this work we propose a new framework to generate deep-learning based condition-aware models that combine information about the type of structure, its various components and the condition of each of the components, all inferred directly from sensed data. Such models can be thought to be analogous to as-built models used in the construction and design industry with automatically generated annotations. Condition-aware models can be used as the basis to obtain higher level assessments on the structure using the statistics and type of the defects observed.

A fully convolutional network (FCN) [16] with a residual network [17] architecture is employed for inference of both building context and damage. A manually labeled dataset of images is created to train each of these networks, which are evaluated for pixel-wise accuracy. The results demonstrate the effectiveness of the proposed trained deep learning models. Finally, a test flight is conducted in Mexico City around a building that was damaged after the 2017 Central Mexico Earthquake to acquire multi-view image data. The UAV survey data is used to build a 3D mesh model onto which the deep learning inference labels are projected to generate a condition-aware model. Section 2 of this paper provides details of the proposed framework, section 3 provides results and section 4 provides the conclusions of the current work.

## 2. Deep learning-based condition aware models

We define condition-aware models as 3D representations of a structure with semantic annotations about the current condition of the structure including but not limited to visual defects such as cracks and spalling and the structure context like type and components. This section details the proposed framework for automatic generation of deep learning-based condition-aware models for rapid post-earthquake inspections.

### 2.1. Framework for generating condition-aware models

The first step of the proposed framework involves image data acquisition of the structure using a UAV to cover as much of the structure as feasible. Next, the acquired images are used to create a 3D mesh model of the structure using the Structure-From-Motion [18]. Deep-learning inference is also conducted on the same set of images to semantically segment (i) damage and (ii) context. The predicted semantic labels are then projected onto the mesh using a UV mapping while averaging overlapping labels from different images to yield a condition-aware model. The resulting model has the 3D mesh structure where each cell has a corresponding averaged label for both damage and context. Figure 1 illustrates the overall framework for the generation of condition-aware models for rapid and automated post-disaster assessments.

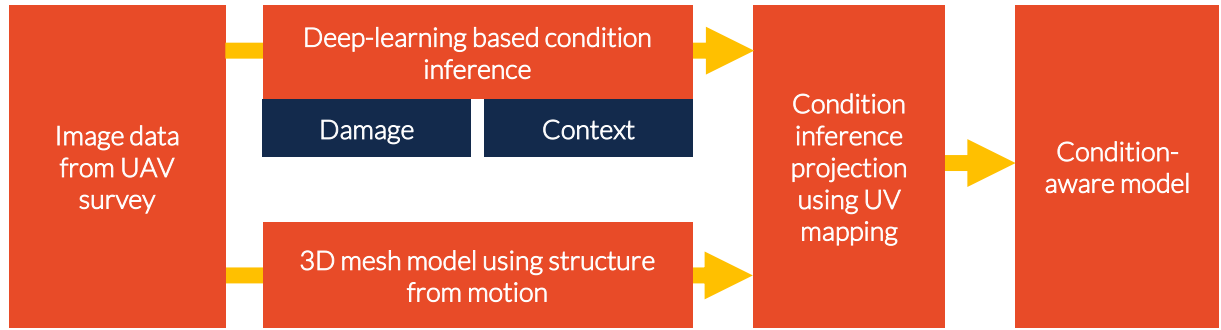

Figure 1. Framework for generation of condition-aware models

### 2.2. Deep learning-based condition inference

Deep learning-based condition inference allows for the automated annotation of damage and context. We propose the use of FCNs for multi-class semantic segmentation of damage and building components to obtain pixel-level information from images. FCNs are neural networks with each layer representable in the form of a convolution [16], i.e.,

$$y_{ij} = f_{ks}(\{x_{si+\delta i, sj+\delta j}\}_{\delta j, \delta i \in 0,k}) \qquad (2.1)$$

where $y_{ij}$ represents the output of any layer, $x_{ij}$ represents the input, $k$ is the kernel size and $f_{ks}$ represents the layer type and could be matrix multiplication, spatial max or elementwise non-linearity. In addition, 'skip' layers are employed to fuse information learnt in lower layers with that learnt in higher ones to increase the fineness of the segmentation. A ResNet architecture with 46 layers is used for each of the networks. The details of the proposed network architecture is shown in Figure 2. Residual connections are used between alternate layers (e.g.,

Conv0 to Conv2, Conv2 to Conv4, etc.). A rectified linear unit is used as the non-linearity for all layers of the network.

Three networks are trained in parallel. A schematic illustration of the three networks is provided in Figure 3. The first network is for scene and building information (scene and building - SB network), second is for identifying the presence of damage (damage presence - DP network) and the third is to identify the type of damage (damage type - DT network). The SB removes any nuisance false positive identifications of damage that may occur outside the building. Awareness of different building components will be necessary to make reasonable assessments and so this network is also trained to identify window/door pixels in the image. The DP network produces a binary classification of damage or no-damage for every pixel. The requirement for the DP network was identified empirically, to reduce the false positive rates of large area damage such as [13].

| Layer | Size | Layer | Size |
| --- | --- | --- | --- |
| Conv0 | 7x7x64 (stride 2) | Maxpool1 | 2x2 |
| Conv1-8 | 3x3x64 | Conv21-32 | 3x3x128 |
| Maxpool0 | 2x2 | Maxpool2 | 2x2 |
| Conv9-20 | 3x3x64 | Conv33-44 | 3x3x128 |

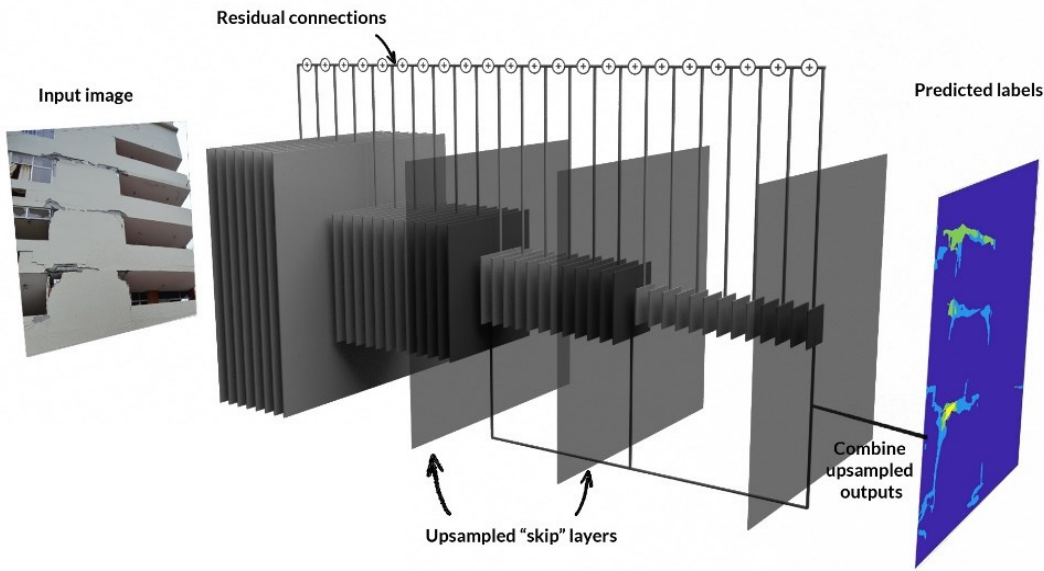

Figure 2. Proposed network architecture. Top: Network layer details. Bottom: Pictorial representation of network.

### 2.3. Datasets used for model training

The SB network was trained on a dataset created with images available from different post-disaster reconnaissance surveys on datacenterhub.org [19]. The dataset included 1000 images of size 288x288. All of the images were hand labelled pixel-wise using a custom-made

Matlab GUI. Eight different classes were labelled including building, wall/door, footpath/pavement, debris, sky, tree, person and vehicle. The DP and DT networks were trained on a subset of the dataset of damaged structures proposed in Hoskere et al. [13]. A total of 665 images of size 288x288 - including damage states relevant to post-earthquake scenario such as concrete cracks, spalling and exposed rebar - were used in the present study. The information from the three networks was combined using a manual softmax threshold and some simple heuristics; for example, spalling on trees, exposed rebar on windows/doors, cracks in the sky, etc., were all ignored, if found. Each network was trained with 80% of the images for 3000 iterations with progressively decreasing learning rates (1e-3 for the first 1500 iterations, 1e-4 for the next 750 iterations, 1e-5 for the next 500 iterations, and 1e-6 for the last 250 iterations). The remaining 20% of the data was used for evaluation. The training was conducted in Tensorflow [20] using an Adam optimizer [21].

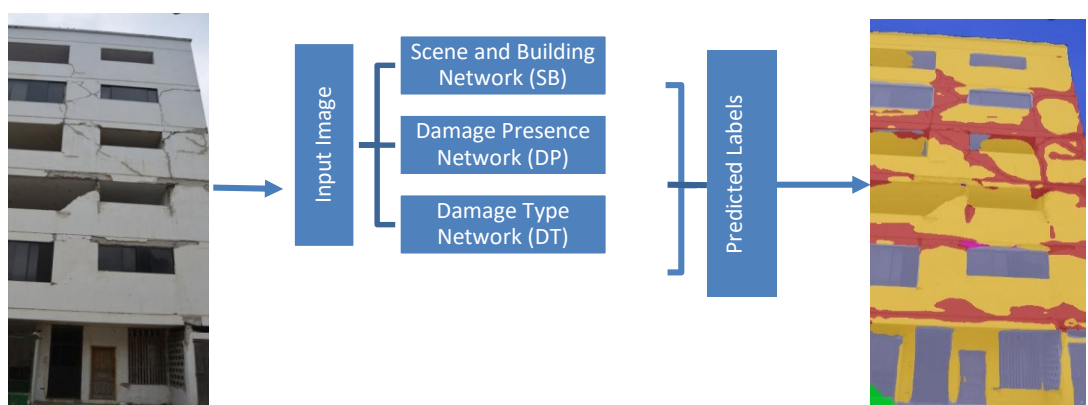

Figure 3. Schematic illustration of the three networks used for deep learning inference

## 2.4. 3D Mesh model using structure from motion

A 3D mesh model of the structure is created using a multi-view stereo workflow. Point correspondences between images are identified using the SIFT (Scale invariant feature transform) feature matching technique [22]. Triangulation is then conducted between corresponding points to generate a dense point cloud. Finally, a dense surface reconstruction is conducted using a Poisson surface reconstruction [23] which allows the use of locally supported basis functions that reduces the solution to a well-conditioned sparse linear system.

## 2.5. Condition inference projection using UV mapping

The final step is to map the predicted labels back on to the 3D surface. Before mapping the labels, a texture atlas has to be created that maps the image coordinates to the 3D mesh coordinates. The process of assigning an object coordinate in 3D space (x, y, z) to a 2D texture coordinate (u, v) is known as UV mapping [24]. The images used to make the 3D mesh of the structure are intelligently blended to create a texture atlas by parameterizing the surface [25] resulting in a UV map. Finally, the same UV mapping is used to map the labels to the 3D surface, thus associating each mesh cell with a set of labels which are then averaged to produce the final labeling.

## 3. Experiments and Results

The framework was evaluated in two stages. First each of the deep networks used for condition inference were evaluated using pixel-wise accuracy and qualitative examination. Next, the overall framework was qualitatively evaluated on a new dataset collected after the 2017 Central Mexico Earthquake.

### 3.1. Demonstration of condition-aware model framework

After the 2017 Central Mexico Earthquake at least 1997 buildings showed significant signs of damage according to the CICM inspection reports[2][1]. Out of these, 1210 were declared low risk, 460 were declared high risk and 327 were tagged as undetermined risk. An elevation survey of one heavily damaged building in the Sona Rosa district of Mexico City was surveyed using a Phantom 3 Pro drone. The overall framework proposed is demonstrated using data collected after the earthquake damaged this building. 250 images from the drone survey were used to generate a 3D surface model using the workflow described in section 2.3 including the generation of texture atlas and UV mapping (Figure 4a). The workflow was implemented using Agisoft Photoscan [25]. The same images were run through the trained deep networks to produce the damage and context labels. A sample result is shown in Figure 4b. The generated UV map was then used to project the labels onto the generated 3D textured model as shown in Figure 4c. The various defects have been highlighted with colored masks where the intensity of each color represents the number of source images that render the particular label.

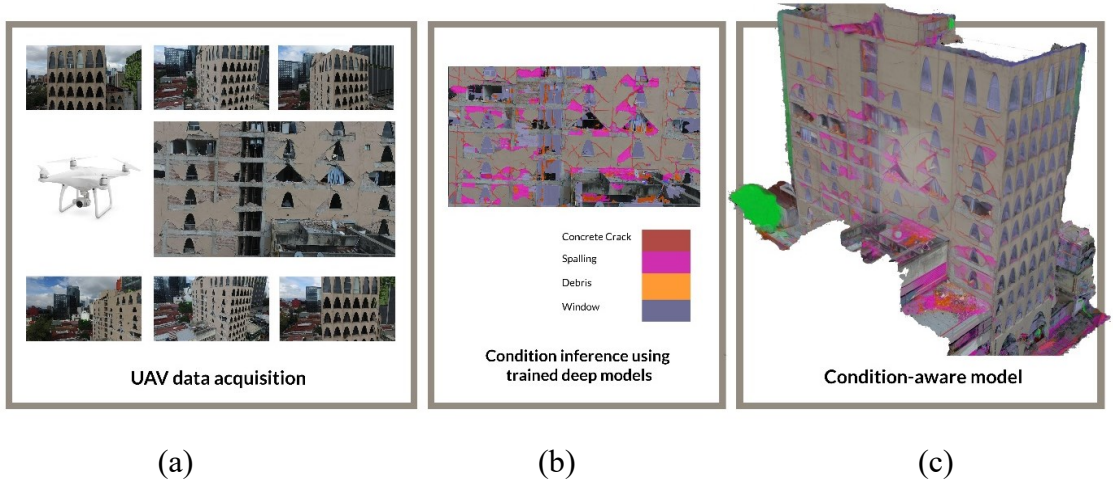

(a)        (b)        (c)

Figure 4. Sample images and generated condition-aware model from survey from Mexico City earthquake

### 3.2. Evaluation of deep learning-based condition inference

Qualitative results for several images are shown in Figure 3, where the right column provides comments regarding the successes, as well as false positives and negatives. Quantitative results are given in Figure 6. The SB network produces good results with pixel-wise accuracies above 80% for each of the eight different classes and an average accuracy of 88.8%. The DT network has a high false positive rate, but the addition of the DP network reduces the false positive rate by 14%, with the combined net having an average accuracy of 91.1%. Using two separate networks for damage improves results by increasing the data efficiency, i.e., the networks are able to be trained with smaller amounts of data. These results demonstrate the effectiveness of the proposed method.

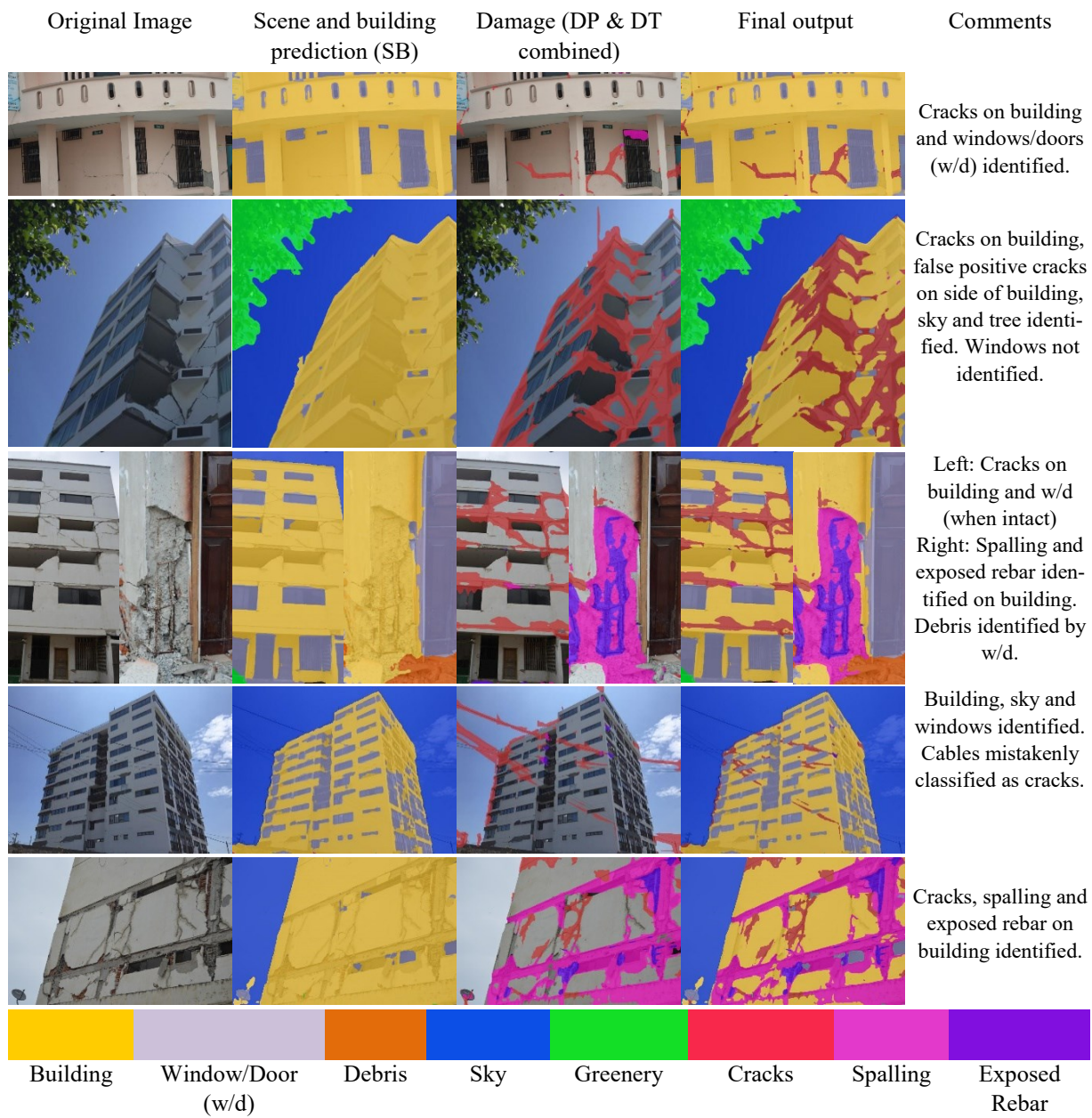

Figure 5. Qualitative results from the proposed network.

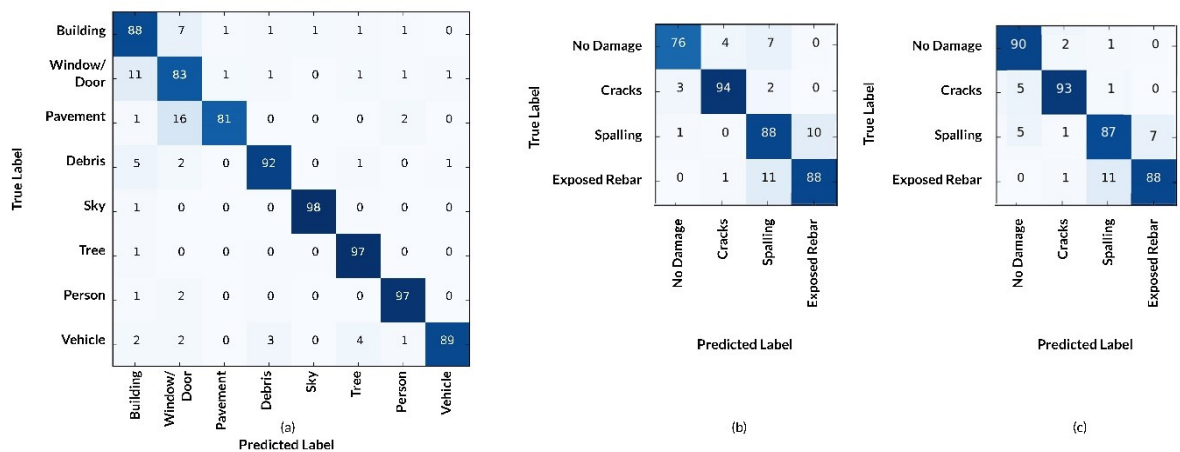

Figure 6. Confusion matrices (a) SB network (b) DT network (c) DP & DT networks combined output.

## 4. Conclusions

Automated post processing to convert UAV data into actionable information is a key step in enabling automated post-disaster inspections. A new framework to generate condition-aware models was proposed to help automate and speed up post-disaster inspections. The framework includes 3D model generation and deep learning-based condition inference of context and damage. Three different networks were used - one for scene and building information (SB), one to identify the presence of damage (DP), and another to identify the type of damage (DT). The SB network had an average accuracy of 88.8% and the combined DP & DT networks together had an average accuracy of 91.1%. The proposed method was successfully able to identify the location and type of damage together with some context about the scene and building. The authors hypothesize that a number of such networks, trained to learn different but related knowledge together with geometric information, will provide enough context to be deployed for automated or semi-automated post-disaster assessments. The generation of condition aware models was demonstrated on a new dataset of images obtained from a damaged structure in Mexico City.

## 5. Acknowledgement

The authors would like to acknowledge funding from the Will K. Brown endowment from the Department of Civil Engineering at the University of Illinois for funding the data collection in Mexico City. The authors would like to acknowledge the help of the members of the Smart Structures Technology lab in labelling the photos, and extend their deep gratitude to Hao Zhou, Siang Zhou, Xintao Wang, Peisong Wu, Xu Chen, Guangpan Zhou, Fernando Gomez, Li Zhu, Xinxia Li, Dongyu Zhang and Yuguang Fu. The authors would also like to thank Michael J. Neal and Prof. Manuel Ruiz-Sandoval for the help during the data collection in Mexico City.